\pdfoutput=1
\documentclass[11pt]{article}

\usepackage{acl}

\usepackage{times}
\usepackage{latexsym}

\usepackage{tcolorbox}

\usepackage[T1]{fontenc}
\usepackage[utf8]{inputenc}
\usepackage{amsmath}

\usepackage{microtype}

\usepackage{inconsolata}

\usepackage{graphicx}

\usepackage{multirow}
\usepackage{booktabs}

\usepackage{subcaption}

\title{Fact-Aware Multimodal Retrieval Augmentation for Accurate Medical
Radiology Report Generation}

\author{Liwen Sun$^*$, James Zhao$^*$, Megan Han, Chenyan Xiong \\
       School of Computer Science, Carnegie Mellon University\\
       \texttt{\{liwens,jjzhao2,wenjingh,cx\}@andrew.cmu.edu}
       }

\definecolor{midnightgreen}{rgb}{0.0, 0.29, 0.33}

\begin{document}
\maketitle

\begin{abstract}
\label{sec:abstract}
%
% Chest radiology is critical for diagnosing cardiac diseases. yet the manual generation of radiology reports is both time-consuming and prone to errors.
%
Multimodal foundation models hold significant potential for automating radiology report generation, thereby assisting clinicians in diagnosing cardiac diseases.
However, generated reports often suffer from serious factual inaccuracy.
%
% Retrieval-augmented generation has emerged as a solid solution to alleviate this issue by grounding report generation with high-quality retrieved reports using the query radiology image.
% %
% However, current medical multimodal retrievers only retrieve reports in a coarse-grained manner, guided solely by radiology diagnostic labels.
In this paper, we introduce a fact-aware multimodal retrieval-augmented pipeline in generating accurate radiology reports (FactMM-RAG).
We first leverage RadGraph to mine factual report pairs, then integrate factual knowledge to train a universal multimodal retriever. 
Given a radiology image, our retriever can identify high-quality reference reports to augment multimodal foundation models, thus enhancing the factual completeness and correctness of report generation.
Experiments on two benchmark datasets show that our multimodal retriever outperforms state-of-the-art retrievers on both language generation and radiology-specific metrics, up to 6.5\% and 2\% score in F1CheXbert and F1RadGraph.
Further analysis indicates that employing our factually-informed training strategy imposes an effective supervision signal, without relying on explicit diagnostic label guidance, and successfully propagates fact-aware capabilities from the multimodal retriever to the multimodal foundation model in radiology report generation.\footnote{Our code is available at \url{https://github.com/cxcscmu/FactMM-RAG}}\def\thefootnote{*}\footnotetext{These authors contributed equally to this work.}

\end{abstract}

\section{Introduction } 
\label{sec:intro}
Within hospitals worldwide, chest radiology serves as a critical technique in identifying cardiac diseases and abnormalities. 
Results of a chest radiograph are typically consolidated in a radiology report, including the source X-ray and a radiologist-produced findings section detailing clinical observations. 
Manually generating these reports, however, can be both time-consuming and potentially inaccessible in under-resourced hospitals \citep{speets2006,iyeke2022}.
Recent multimodal foundation models have exhibited remarkable capabilities in challenging healthcare tasks, motivating an automation of this process to enhance physicians' efficiency on clinical decision-making and improve patient health outcomes \citep{_all__2021,li2023llavamed,  pmlr-v225-moor23a, tu2023generalist,sun2024edcopilot}.\\
\newline
Although prior medical multimodal foundation models have demonstrated promising capabilities on report generation given the radiology image, they still suffer from serious hallucinations by generating factually inaccurate reports \citep{pal2023medhalt,ahmad2023creating,pal2024gemini}.
Factual correctness is especially critical in chest radiology domains, as minute textual differences can drastically invert radiology report meaning and downstream prescribed treatments \citep{delbrouck-etal-2022-improving,xie2023factreranker,liu2024factual}.
Retrieval-Augmented Generation (RAG) has emerged as a popular paradigm to address this issue by grounding text generation with retrieved relevant knowledge given a query \citep{lewis2021retrievalaugmented,chen2022murag,gao2024retrievalaugmented}.
However, developing medical multimodal retrievers remains challenging, requiring retrievers to bridge the gap between symptomatic image semantics and factually-equivalent report text.\\
\newline
To capture fine-grained details in chest radiographs and improve the factual completeness of generated reports, we introduce FactMM-RAG, a fact-aware multimodal retrieval-augmented pipeline for generating accurate radiology reports given a radiology image.
By designing a novel report pair-mining procedure incorporating factual knowledge, we develop a fact-aware retriever to augment multimodal foundation models in generating accurate chest X-ray radiology reports. 
Specifically, we first leverage RadGraph \citep{jain2021radgraph} to mine factually-oriented report pairs by annotating consistent radiology entities and relations between query and reference reports with certain abnormalities. 
Next, we train a universal multimodal encoding architecture through mined report pairs to conduct multimodal dense retrieval. 
Given an unseen patient's radiology image, our retriever encodes it and searches for the most similar factually-informed reference report from an available report corpus.
Passing them together into a multimodal foundation model unlocks its fact-aware potential to generate more accurate radiology reports.\\
\newline
Our experiments reveal that our retriever outperforms all state-of-the-art retrievers in both language generation and clinically relevant metrics on the MIMIC-CXR and CheXpert datasets, achieving up to 6.5\% and 2\% score in F1CheXbert and F1RadGraph for final RAG evaluation.
We also investigate our retriever's fact-aware capability controlled by factual similarity thresholds and confirm that our factually-informed training strategy can impose a useful supervision signal without relying on explicit diagnostic label guidance.
Further analysis through retrieval evaluation metrics shows that the fact-aware capability of our retriever can be effectively propagated to the multimodal foundation models.
Lastly, our case study highlights that among reports describing the same symptom from different retrievers, those generated by our model are more accurate and achieve greater factual correctness.\\
\newline
Our main contributions can be summarized as follows:
\begin{itemize}
    \item We propose a fact-aware medical multimodal retriever to augment multimodal foundation models in generating accurate chest X-ray radiology reports.
    \item We design a method for mining factually-informed radiology report pairs that trains multimodal encoders to retrieve high-quality reference reports.
    \item We demonstrate that on two benchmark datasets, our medical multimodal retriever outperforms state-of-the-art medical multimodal retrievers on both language generation and clinically relevant metrics.
\end{itemize}
The rest of this paper is organized as follows.
We review related work in in Section \ref{sec:related}. We discuss the pipeline of FactMM-RAG in Sections \ref{sec:method}. Section \ref{sec:setup} and \ref{sec:results} discuss our experimental setup and results.

\section{Related Work}
\label{sec:related}
\begin{figure*}[t]
    \centering
    \includegraphics[width=\textwidth]{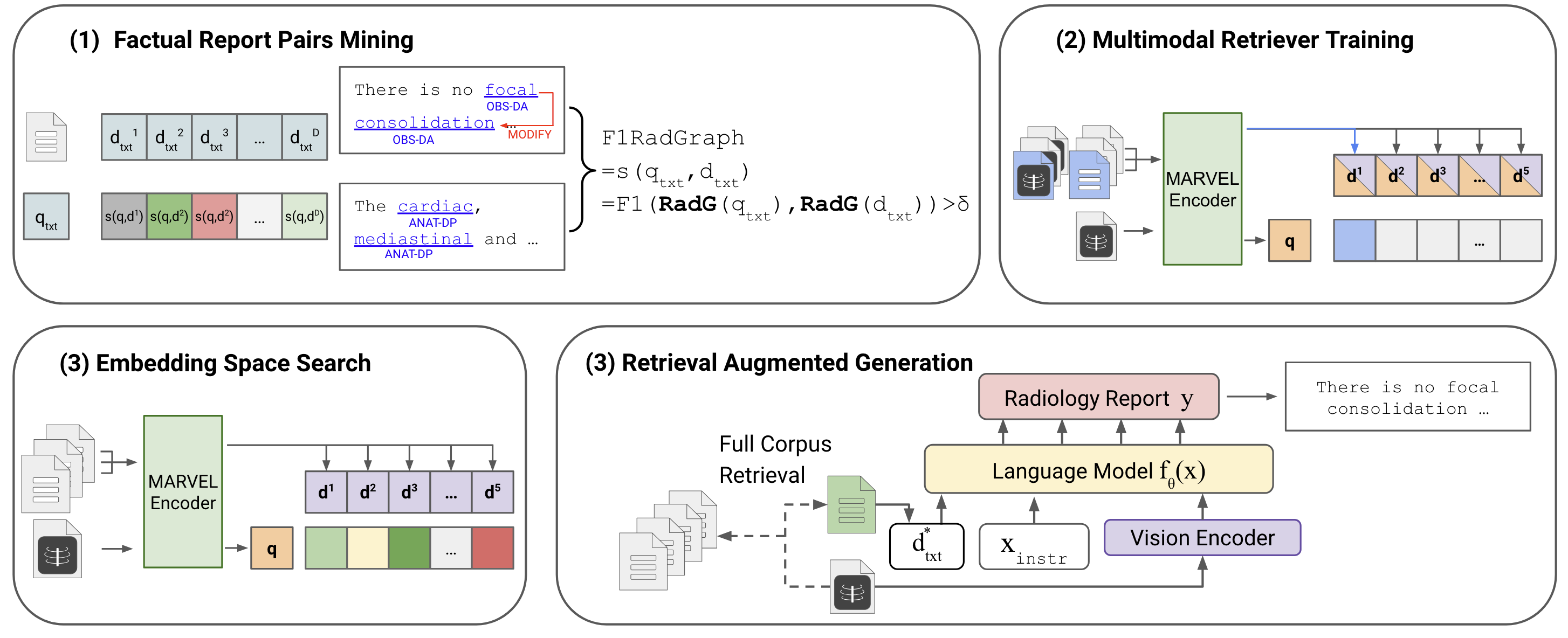}
    \caption{An overview of the FactMM-RAG system. It mainly contains three stages: (1) Leveraging RadGraph to characterize each radiology report and mine factually-informed report pairs; (2) Integrating factual knowledge into the training of the universal multimodal retriever; (3) Given the radiology image, employing the fact-aware multimodal retriever to search for factually-informed reference reports and augmenting the multimodal foundation model in generating accurate radiology reports.}
    \label{fig:system_digram}
\end{figure*}
\textbf{Retrieval Augmented Generation.} Retrieval Augmented Generation, utilizing external knowledge to enhance language models, has shown great promise in text-generation performance on factual accuracy especially for Open-Domain QA. \citep{borgeaud2022improving,izacard2022atlas}. 
\citet{guu2020realm,lewis2021retrievalaugmented} involve end-to-end training through both generators and retrievers;  
\citet{shi2023replug,yu2023augmentationadapted} adapt the end-to-end pattern by employing black-box LLM training signal propagation for retriever tuning. 
%
% \citet{lin2024radit}, in particular, jointly leverages retriever fine-tuning from LLM preferences, alongside LLM retrieved-document utilization.\\ 
%
% \newline
%
Further works have expanded RAG to multiple modalities, employing unified image-text encoders \citep{radford2021learning} or separate pretrained encoders \citep{dosovitskiy2021image,raffel2023exploring} and plugging retrieved documents into multimodal foundation models \citep{chen2022murag,hu2023reveal}.  
\citet{yasunaga2023retrievalaugmented} similarly integrates multimodal retrieval with both text and image generation capabilities.\\
\newline
\textbf{Medical Multimodal Retriever.} 
Joint training of image-text pairs in a shared embedding space, as exemplified by CLIP \citep{radford2021learning}, facilitates visual and textual modality interactions, providing flexible representations for general-domain downstream tasks.
Adapting general-domain multimodal retrievers to medical domains, however, is non-trivial due to the necessity of specialized knowledge.
\citet{zhang2022contrastive} introduces an unsupervised approach for radiology image representation learning from paired text descriptions.
\citet{huang2021gloria} leverages global image-report and local sub-region features for multimodal retrieval and classification. 
\citet{wang2022medclip,You_2023} propose medical knowledge extraction for constructing contrastive learning image-text pairs. 
\citet{zhang2024biomedclip} addresses the limited diversity within medical datasets, curating a large biomedical image-text collection towards a biomedical multimodal foundation model.
Nevertheless, these existing medical multimodal retrievers neglect specific image information and do not adequately emphasize factual accuracy, resulting in imprecision when retrieving radiology reports.\\
\newline
\textbf{Medical Multimodal Foundation Model.} 
Significant efforts have been made in applying multimodal foundation models to the medical imaging domain \citep{li2023llavamed,  pmlr-v225-moor23a, tu2023generalist,sun2024edcopilot}. 
As chest X-ray radiology is the most commonly performed imaging examination, tailored medical multimodal foundation models for this critical area has gathered much attention \citep{ chambon2022roentgen, chen-etal-2021-cross-modal, Omkar2023XrayGPT, wu2023generalist, chen2024chexagent}.
\citet{jain2021radgraph} advances this area by designing a novel information extraction schema to structure radiology reports from chest radiographs;
\citet{miura2021improving, delbrouck-etal-2022-improving} take a step forward, using reinforcement learning from semantic rewards to improve the factual quality of generated radiology reports;
\citet{chen2024chexagent} recently has also developed an instruction-tuned multimodal foundation model capable of sophisticated interpretation and analysis of chest X-rays.\\
%
% \cx{let's discuss very clearly that our focus is the report generation with the image only, rather than doing the summarization task which is the step after report generation.}
\newline
% \cx{we need more detail on the radiology report generation related work, as it is the most related one.}
%
One closely related line of work to ours is retrieval-based radiology report generation given only radiology images.
For instance, \citet{li2018hybrid} proposes a retrieval policy module to update radiology reports via hierarchical reinforcement learning;
\citet{pmlr-v158-endo21a} employs image-text embeddings from contrastive learning for retrieval-augmented radiology report generation;
\citet{ramesh2022improving} proposes synthesizing additional reports and reducing hallucinations from reference report priors to improve medical radiological report generation.
\section{Methodology }
\label{sec:method}
%
% jz - working on diagrams here
% https://docs.google.com/presentation/d/1eW_wDYQh6ITt_gj0DzcEJ3I2heeiMv81xEoA8CbpZGs/edit?usp=sharing
%
In this section, we present the overall methodology of FactMM-RAG.
We first detail the training procedure of our fact-aware medical multimodal retriever in Section \ref{subsec:retriever}.
We then provide the pipeline for retrieval-augmented radiology report generation with our multimodal retriever in section  \ref{subsec:rag}. The overview is illustrated in Figure \ref{fig:system_digram}.

\subsection{Fact-aware Multimodal Retrieval}
\label{subsec:retriever}
This section discusses the training process of the multimodal retriever with factual knowledge. Each patient in the corpus has a chest X-ray radiology image along with its corresponding report.
% \cx{I prefere we describe our method in a more mathmatically way.}
%
We begin by annotating each report using RadGraph \citep{jain2021radgraph}, then constructing factual report pairs to train our multimodal retriever.
We describe these steps as follows.\\
\newline
\textbf{Chest Radiograph Annotation.} 
Since radiology reports are free-text, we utilize the RadGraph information extraction tool to extract structured knowledge graphs from them.
Specifically, RadGraph employs named entity recognition and relation extraction models to identify radiological entities (e.g. carina, lungs, abnormalities) and the clinical relations between them (e.g. modify, located at, suggestive of).
Each radiology report is then segmented into distinct regions and stored as $[(\texttt{entity}_{1}, \texttt{entity label}_{1}, \texttt{relation}_{1}), (\texttt{entity}_{2},\\
\texttt{entity label}_{2}, \texttt{relation}_{2}), \ldots]$.
After characterizing the chest radiograph for each report in the training corpus, we construct factual report pairs.
\\
\newline
\textbf{Factual Report Pairs Mining.}
Each report has an associated medical label describing the symptom.
We first utilize the query report to search for other reports with the same symptom,  aiming to eliminate false negatives when constructing report pairs.
Rather than solely relying on the diagnostic labels, we further capture the factually-oriented pathology semantics between different reports. Following F1RadGraph \citep{jain2021radgraph},
we calculate the factual similarity $s(q_{txt},d_{txt})$ between query report $q_{txt}$ and other reports $d_{txt}$ in the annotated format as follows,
\begin{align}
\label{eq:radgraph_similarity}
s(q_{txt},d_{txt}) = \frac{2\cdot ( \hat{q}_{txt} \cap  \hat{d}_{txt})}{\text{length}(\hat{q}_{txt})+\text{length}(\hat{d}_{txt})},
\end{align}
where $\hat{q}_{txt},\hat{d}_{txt}$ denotes reports with only annotated entities and relations in RadGraph structured form. We then set a strict threshold $\delta$ to filter out searched reports with low similarity score:
\begin{align}
\label{eq:threshold}
N_{q_{txt}} = \{ d_{txt} \in D |   s(q_{txt},d_{txt}) > \delta \}.
\end{align}
where $N_{q_{txt}}$ denotes factual positive report pairs for $q_{txt}$ and $D$ is the total training corpus. 
Since each query report is associated with a corresponding radiology image, these factual report pairs can also be applied to the query report's radiology image.
Next, we train our multimodal retriever with mined factual report pairs.\\
\newline
\textbf{Multimodal Dense Retrieval.} Following previous work \citep{zhou2024marvel}, we universally encode each query  image $q_{img}$ and other  image-text pairs $(d_{txt},d_{img})$ in the training corpus, using one encoder, MARVEL:
\begin{align}
    \mathbf{q} &= \text{MARVEL}(q_{img});\\
    \mathbf{d} &= \text{MARVEL}(d_{txt},d_{img}),
\end{align}
where each image-text pair is represented as a single embedding. 
We then model the relevance score $f(q,d)$ between the query image and other image-text pairs by cosine similarity:
\begin{align}
    f(q, d) = \cos(\mathbf{q}, \mathbf{d}).
\end{align}
To inject factually-oriented medical knowledge into multimodal retrieval, we train the encoder to minimize the following loss, 
%\scalebox{1.5}
\begin{equation}
\tiny
\mathcal{L}= -\sum_{q_{img} \in D} \sum_{d^+ \in N_{q_{img}}}\log \frac{e^{f(\mathbf{q}, \mathbf{d}^+) / \tau}}{e^{f(\mathbf{q}, \mathbf{d}^+) / \tau} + \sum_{\mathbf{d}^-} e^{f(\mathbf{q}, \mathbf{d}^-) / \tau}},
% &\propto - \underbrace{f(q, d^+) / \tau}_{L_{\text{Align}}} + \log ( \sum_{d^- \in \mathcal{D}^-} ( \underbrace{e^{f(q, d^-_{\text{Image}}) / \tau}}_{L_{\text{Image}}} + \underbrace{e^{f(q, d^-_{\text{Text}}) / \tau}}_{L_{\text{Text}}} )),
\end{equation}
where $d^+$ are obtained through factual report pair mining and $d^-$ are in-batch negative samples \citep{karpukhin2020dense}.
Then, we use our multimodal retriever and foundation model to perform retrieval-augmented radiology finding generation.
\subsection{Retrieval Augmentation for Accurate Radiology Report Generation}
\label{subsec:rag}
Given our trained fact-aware multimodal retriever, we encode the query image and each report in the training corpus.
Then, we retrieve the report with the highest relevance score to the query image as the factually-informed relevant report.
Subsequently, we pass the query image along with the relevant report into a multimodal foundation model to perform retrieval-augmented generation training.
The multimodal foundation model is finetuned by standard autogressive loss,
\begin{equation}
    \small
    \mathcal{L} = -\frac{1}{n}\log \prod_i^n p_\theta(y_i|q_{img}, d_{txt}^*, x_{\texttt{instr}}, y_{<i}), 
\end{equation}
where $q_{img}$ is the query image, $d_{txt}^*$ is the retrieved factually-informed relevant report, $x_{\texttt{instr}}$ is the prompt instruction, and $y$ is the ground-truth report.
During inference, we retrieve a relevant report from the training corpus using an unseen patient X-ray image, and pass them into the multimodal foundation model to generate findings with higher factual accuracy.
% we use the nearest-neighbor's corresponding finding $d_{txt}^j$ as a plug-in to perform standard Retrieval Augmented Generation. We follow LLaVA \citep{liu2023visual} in using a single-turn conversation $(x_{q}^1, x_{a}^1)$ for an image $q_{img}$, with $x_{\texttt{instr}}^1=[q_{img}, x_{q}^1]$ . For Retrieval-Augmented Generation, we embed the retrieved document $d_{txt}^j$ within the initial question $x_q^1$, with prompting details enumerated in the appendix. The model's answers are fine-tuned using autoregressive Cross-Entropy Loss. At test time, the unseen patient's x-ray embedding is searched against a corpus of X-ray/finding pairs, in which the closest retrieved corpus document is used in RAG. 

\section{Experimental Setup}
\label{sec:setup}
\textbf{Dataset.} Following \citet{delbrouck-etal-2023-overview}, we use the processed MIMIC-CXR \cite{Johnson2019} to train both retriever and foundation model. 
This dataset contains 125,417 training radiology image-report pairs, 991 validation pairs, and 1,624 test pairs.
They are sourced from the Beth Israel Deaconess Medical Center. 
CheXpert \cite{irvin2019chexpert} is another chest X-ray dataset from Stanford Health Care. 
Since it contains complete finding reports only for a testing dataset containing 1000 pairs, we use it as zero-shot evaluation.
\\
\newline
\textbf{Evaluation Metrics.} We evaluate our proposed system using both natural language generation and medically-tailored evaluation metrics.
For language fluency measures, we use ROUGE-L \cite{Lin2004ROUGEAP} to evaluate the longest common subsequence overlap between the generated and reference findings, and BERTScore \cite{zhang2020bertscore} to evaluate non-clinical semantic sentence similarity.\\
\newline
For clinical accuracy measures,  we use CheXbert \cite{smit2020chexbert} to generate the ground-truth diagnostic labels for finding reports.
%
% identifying 14 different types of observations. 
%
Following \citet{delbrouck-etal-2023-overview}, we then calculate the F1CheXbert \citep{zhang2020optimizing}, which is the F1-score for 5 observations (Cardiomegaly, Edema, Consolidation, Atelectasis, Pleural Effusion) by comparing the generated report with the reference report's classifications. 
Beyond using the limited diagnostic labels for evaluation, we also adopt F1RadGraph \citep{jain2021radgraph} to measure factual correctness by calculating the overlap in radiological entities and clinical relations between the generated report and the reference report.
See Appendix \ref{subsec:app_eva} for more details.\\
\newline
\textbf{Baselines.}
We mainly compare our retriever with other baselines under  multimodal RAG setting.
We include the following baselines, CLIP \citep{radford2021learning} is a multimodal retriever pretrained from general-domain image-text pairs;
GLoRIA \citep{huang2021gloria} leverages attention-weighted image regions with contextual words to learn localized and global representations for radiology images and reports;
MedCLIP \citep{wang2022medclip} and CXR-CLIP \citep{You_2023} build upon CLIP and utilize diagnostic labels as training signals for learning radiology image and text representations;
BiomedCLIP \citep{zhang2024biomedclip} extends the radiology-specific dataset and pretrains on a larger magnitude of biomedical data to learn multimodal representations;
Med-MARVEL utilizes universal encoder MARVEL \citep{zhou2024marvel} to conduct contrastive learning on each patient's self image-report pair without further training on factual image-report pairs.\\
\newline
We also compare our method with non-RAG approach.
"No Retriever" refers to directly fine-tuning the backbone to generate  reports without retrieval augmentation;
ORGan \citep{hou2023organobservationguidedradiologyreport} first creates an observation plan, then feeds the plan and radiographs to generate the report through tree reasoning mechanism. 
Upper-bound results using an oracle in training corpus with top-1 factual similarity to test query report are presented. More details are in Appendix \ref{subsec:app_eva}.\\
\newline
\textbf{Implementation Details.} In our experiments, we use MARVEL \citep{zhou2024marvel} as our multimodal retriever backbone.
MARVEL is a language model based on T5-ANCE \citep{inproceedings_ance}, trained with modality-balanced hard negatives.
We use LLaVA \cite{liu2023visual} as our multimodal foundation model backbone. 
Since each radiology study contains multiple image views for each patient, we select the frontal view.
We also concatenate the finding and impression sections to form the X-ray report.
To reduce training costs and address factual report pair imbalances, we rerank the retrieved reports by factual similarity and use the top 2 factual report pairs for each query to train our multimodal retriever.
We leave more training details in Appendix \ref{subsec:app_ret} and \ref{subsec:app_reg}. 

\begin{table*}[h]
\centering
\resizebox{\textwidth}{!}{
% jz: copied highlighted results here
\begin{tabular}{lccccccccc}
    \toprule
    & \multicolumn{4}{c}{\textbf{MIMIC-CXR}} & \multicolumn{4}{c}{\textbf{CheXpert}} \\
    \cmidrule(lr){2-5} \cmidrule(lr){6-9}
    \textbf{Model} & \multicolumn{2}{c}{\textbf{Factual Similarity}} & \multicolumn{2}{c}{\textbf{Textual Similarity}} & \multicolumn{2}{c}{\textbf{Factual Similarity}} & \multicolumn{2}{c}{\textbf{Textual Similarity}} \\
    \cmidrule(lr){2-3} \cmidrule(lr){4-5} \cmidrule(lr){6-7} \cmidrule(lr){8-9}
    & F1CheXbert & F1RadGraph & ROUGE-L & BERTScore & F1CheXbert & F1RadGraph & ROUGE-L & BERTScore \\
    \midrule
    No Retriever & 0.496 & 0.234 & 0.294 & 0.549 & 0.371 & 0.173 & 0.231 & 0.469 \\
    ORGan \cite{hou2023organobservationguidedradiologyreport} & 0.541 & 0.240 & 0.308 & 0.552 & 0.431 & 0.181 & 0.232 & 0.470 \\
    \midrule
    CLIP  \cite{radford2021learning} & 0.507 & 0.241 & 0.300 & 0.552 & 0.381 & 0.172 & 0.231 & 0.468 \\
    GLoRIA \cite{huang2021gloria} & 0.476 & 0.232 & 0.294 & 0.543 & 0.397 & 0.173 & 0.231 & 0.468 \\
    MedCLIP \cite{wang2022medclip}& 0.517 & 0.238 & 0.298 & 0.549 & 0.408 & 0.182 & 0.238 & 0.471   \\
    CXR-CLIP \cite{You_2023} & 0.501 & 0.243 & 0.302 & 0.553 & 0.406 & 0.183 & 0.241 & 0.471 \\
    BiomedCLIP \cite{zhang2024biomedclip}& 0.502 & 0.233 & 0.293 & 0.546 & 0.380 & 0.173 & 0.232 & 0.469 \\
    \midrule
    Med-MARVEL \cite{zhou2024marvel}& 0.537 & 0.237 & 0.306 & 0.549 & 0.454 & 0.185 & \textbf{0.243} & 0.472 \\
    FactMM-RAG & \textbf{0.602} & \textbf{0.257} & \textbf{0.307} & \textbf{0.561} & \textbf{0.475} & \textbf{0.185} & 0.236 & \textbf{0.475} \\
    \midrule
    Oracle & 0.972 & 0.523 & 0.495 & 0.677 & 0.951 & 0.384 & 0.350 & 0.548 \\
    \bottomrule
\end{tabular}
}
\caption{Overall performance of FactMM-RAG and baselines under the multimodal retrieval-augmentation setting. Models are evaluated by textual similarity and factual similarity between generated and reference reports. FactMM-RAG outperforms the best baseline with p-value < 0.05. }

\label{tab:main_results}
\end{table*}

 % 0.408 & 0.182 & 0.238 & 0.471 \\
 % 0.406 & 0.183 & 0.241 & 0.471
\begin{figure*}[t]
    % Hyperparameter Search
    \vspace{2mm}
    \begin{subfigure}[t]{0.24\textwidth}
        \centering
        \includegraphics[width=\textwidth]{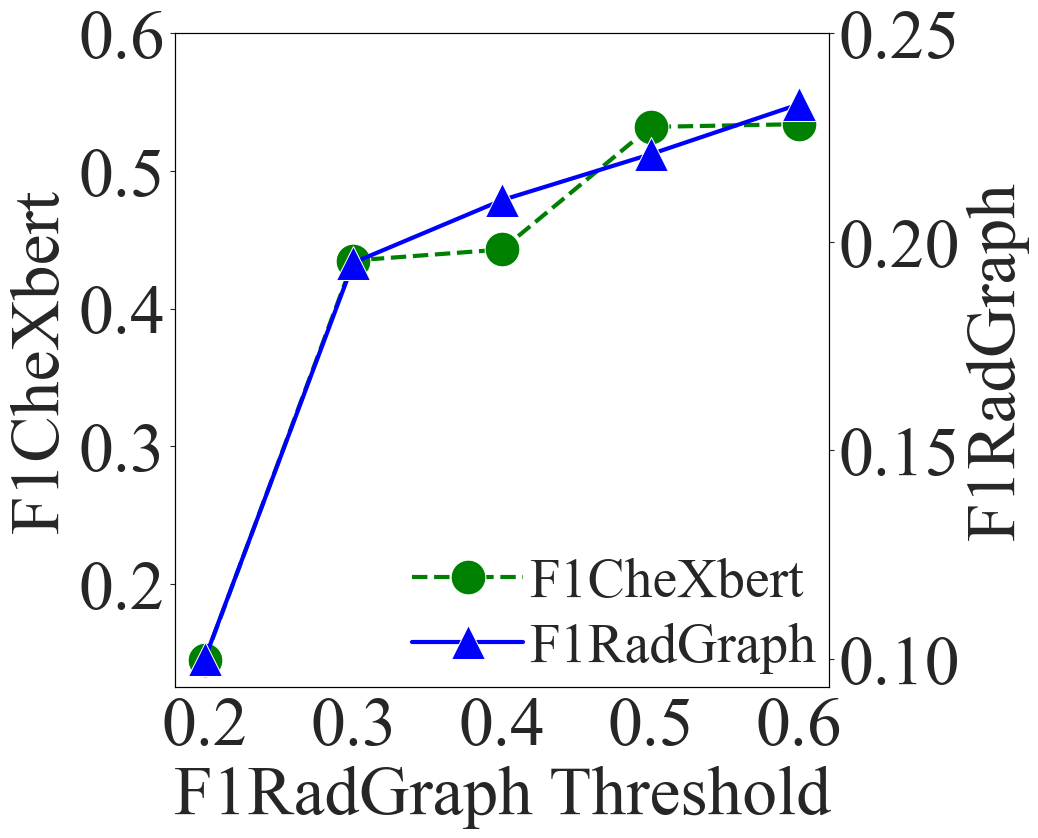}
        \caption{F1CheXbert Threshold: 0.0}
        \label{fig:chex0}
    \end{subfigure}~
    % \hspace{-0.2cm}
    \begin{subfigure}[t]{0.24\textwidth}
        \centering
        \includegraphics[width=\textwidth]{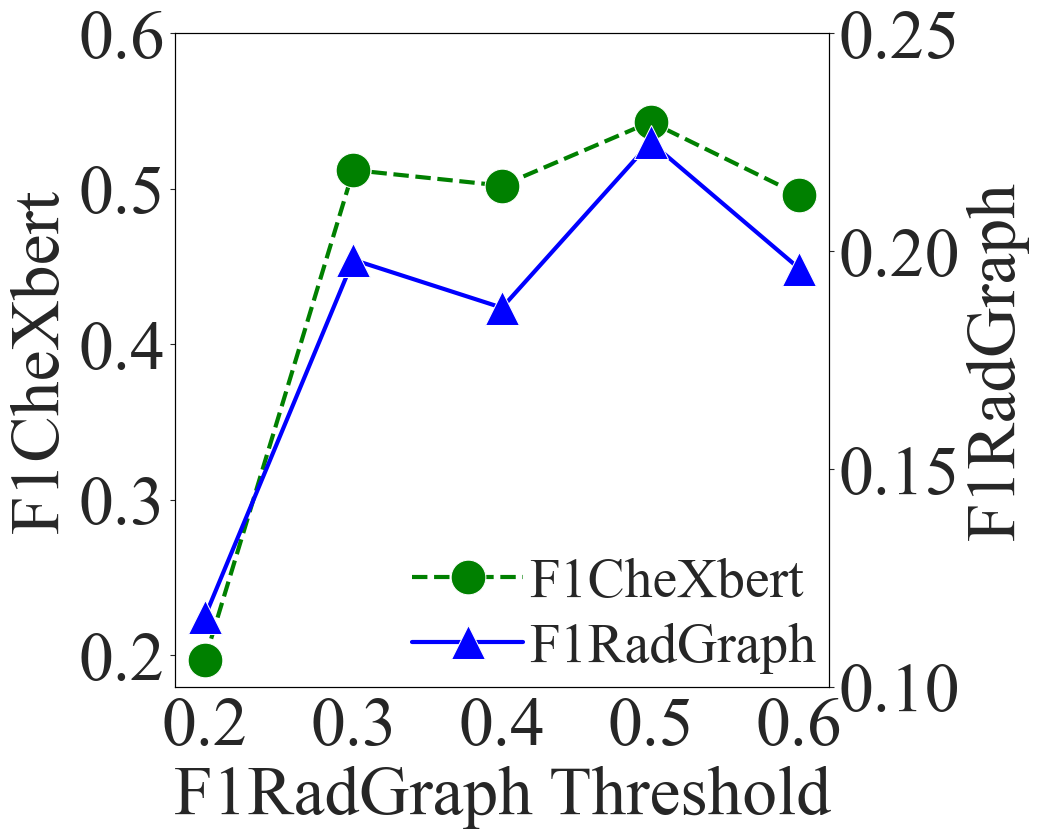}
        \caption{F1CheXbert Threshold: 0.4}
        \label{fig:chex0.4}
    \end{subfigure}
    \begin{subfigure}[t]{0.24\textwidth}
        \centering
        \includegraphics[width=\textwidth]{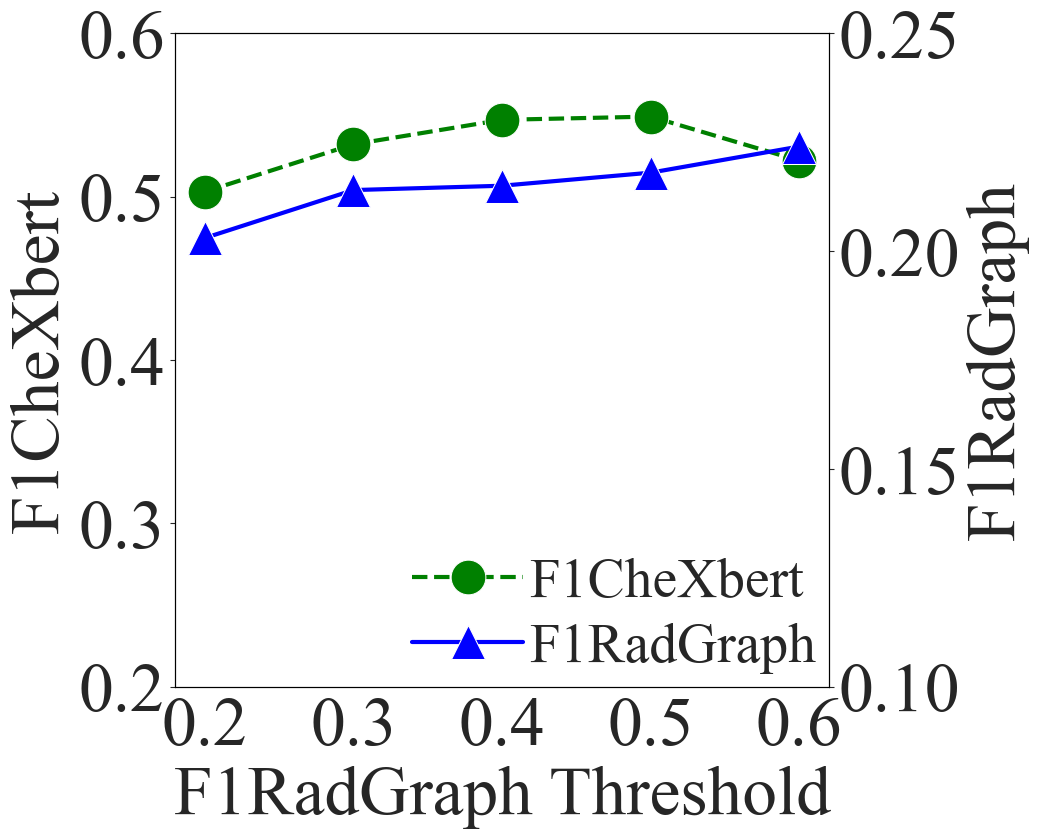}
        \caption{F1CheXbert Threshold: 0.8}
        \label{fig:chex0.8}
    \end{subfigure}
    \begin{subfigure}[t]{0.24\textwidth}
        \centering
        \includegraphics[width=\textwidth]{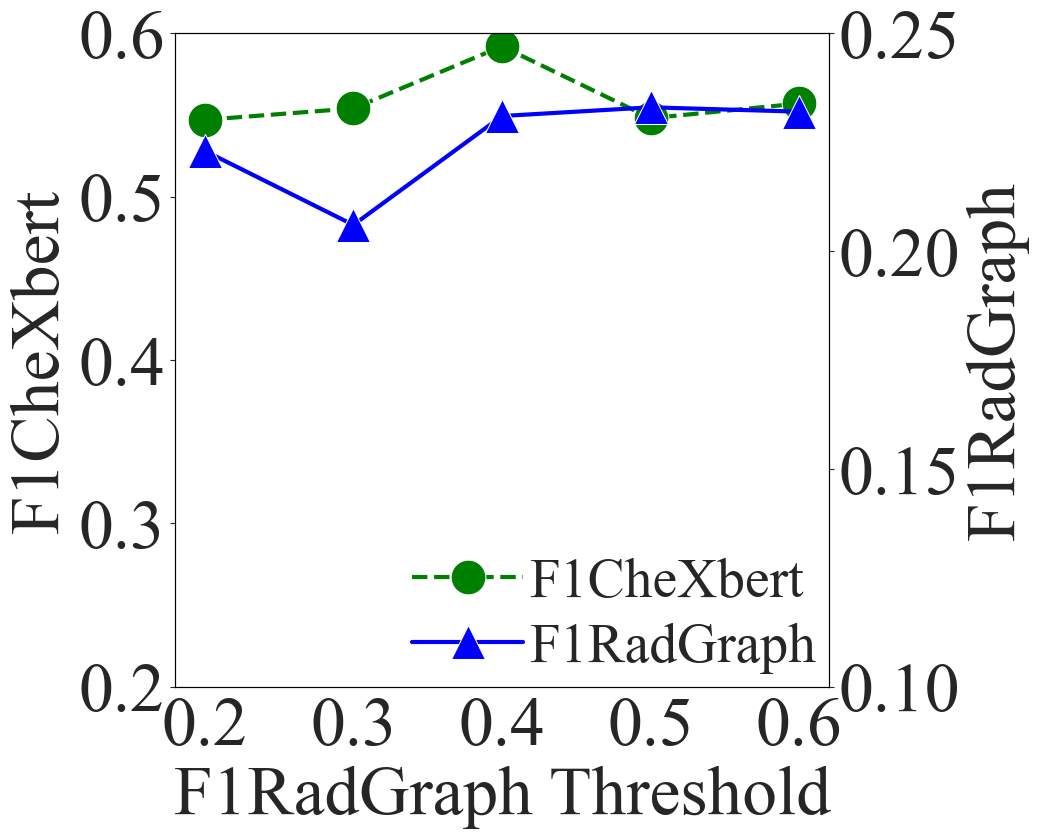}
        \caption{F1CheXbert Threshold: 1}
        \label{fig:chex1.0}
    \end{subfigure}
    % \vspace{-0.3cm}
   \caption{Factual performance of FactMM-RAG controlled by different F1CheXbert and F1RadGraph thresholds. We vary the F1RadGraph thresholds under one fixed F1CheXbert threshold selected from \{0, 0.4, 0.8, 1\}.}
    \label{fig:hyperparameter_search}
\end{figure*}

\section{Evaluation Results}
\label{sec:results}
In this section, we present our experimental results.
We first evaluate the overall performance between different retrievers in section \ref{subsec:overall}. 
Next, we discuss the ablation studies in section \ref{subsec:ablation}. 
We then explore the fact-aware capability of our retriever in section \ref{subsec:threshold} and section \ref{subsec:propa}.
Lastly, we show the superiority of our retriever through a case study in section \ref{subsec:case}.
\subsection{Overall Performance}
\label{subsec:overall}
The results of our fact-aware RAG system are shown in Table \ref{tab:main_results}. 
In MIMIC-CXR, FactMM-RAG outperforms state-of-the-art retrievers by a significant margin, up to 6.5\% in F1CheXbert and 2\% in F1RadGraph.
In the CheXpert zero-shot evaluation, FactMM-RAG outperforms state-of-the-art retrievers by 2\% and 1.2\% in these two metrics, indicating our retriever's generalization capability compared to other models.\\
\newline
To establish the effectiveness of our RAG approach, we also show that FactMM-RAG  significantly outperforms the fine-tuned backbone without retrieval augmentation by 10\% and achieves competitive improvement over the strong non-RAG ORGan baseline. \\
\newline
Besides, we can observe that adopting the baseline retrievers on top of multimodal foundation models only yields marginal gains compared to the finetuning of foundation model generation without retrieval-augmentation. 
This shows that reports retrieved by baseline retrievers are factually-inferior to those from our retriever, potentially passing misleading information that prevents the foundation model from generating factual reports.  \\ 
\newline
Specifically, compared to the retriever Med-MARVEL, we also observe factual-correctness performance gain based on two clinical metrics.
Both use the same universal encoder backbone, but FactMM-RAG benefits from the injected factual medical knowledge, allowing it to search for the most similar and factually correct reports, thereby assisting the multimodal foundation model in generating more accurate reports.
%

% Without, both Med-MARVEL and FactMM-RAG observed improvements across the board. We take this as evidence of the validity of the RAG paradigm as a whole in injecting radiology report generation factual correctness as a whole.  \\
% For final RAG evaluation, \ls{Change it to acronym, RAG}

% med-marvel improves chex w/ maintaning rag
% dpr and ance > everything 

\begin{table*}[t]
\centering

\resizebox{\textwidth}{!}{
% jz: copied highlighted results here
\begin{tabular}{lccccccccc}
    \toprule
    & \multicolumn{4}{c}{\textbf{MIMIC-CXR}} & \multicolumn{4}{c}{\textbf{CheXpert}} \\
    \cmidrule(lr){2-5} \cmidrule(lr){6-9}
    \textbf{Model} & \multicolumn{2}{c}{\textbf{Factual Similarity}} & \multicolumn{2}{c}{\textbf{Textual Similarity}} & \multicolumn{2}{c}{\textbf{Factual Similarity}} & \multicolumn{2}{c}{\textbf{Textual Similarity}} \\
    \cmidrule(lr){2-3} \cmidrule(lr){4-5} \cmidrule(lr){6-7} \cmidrule(lr){8-9}
    & F1CheXbert & F1RadGraph & ROUGE-L & BERTScore & F1CheXbert & F1RadGraph & ROUGE-L & BERTScore \\
    \midrule
    \multicolumn{9}{l} {\textbf{Setting}: Multimodal Retrieval}\\
    \midrule
    CLIP  \cite{radford2021learning} & 0.341 & 0.160   & 0.238 & 0.489  & 0.285 & 0.130   & 0.207 & 0.439 \\
    GLoRIA \cite{huang2021gloria}& 0.346 & 0.137   & 0.211 & 0.453 & 0.359 & 0.135   & 0.216 & 0.447 \\
    MedCLIP \cite{wang2022medclip} & 0.539 & 0.198   & 0.261 & 0.508  & 0.478 & 0.161   & 0.225 & 0.454   \\
    CXR-CLIP \cite{You_2023}& 0.516 & 0.215   & 0.277 & 0.524  & 0.444 & 0.167   & 0.230& 0.458   \\
    BiomedCLIP \cite{zhang2024biomedclip}& 0.502 & 0.233 & 0.293 & 0.546 &  0.386 & 0.142   & 0.216  & 0.441\\
    Med-MARVEL \cite{zhou2024marvel}& 0.550 & 0.212   & 0.279  & 0.525  & 0.479 & 0.160   & 0.222 & 0.454  \\
    FactMM-RAG  & \textbf{0.605 }&\textbf{0.249}   & \textbf{0.297}   & \textbf{0.547}  & \textbf{0.491} & \textbf{0.174}   & \textbf{0.237} & \textbf{0.467 } \\
    \midrule
    Oracle & 0.992 & 0.429  & 0.399  & 0.612 & 0.999 & 0.438 & 0.362 & 0.554 \\
        \midrule
    \multicolumn{9}{l} {\textbf{Setting}: Multimodal Retrieval Augmented Generation}\\
    \midrule
     ClueWeb-LLaVA$_{1.5}$  & \textbf{0.602} & 0.257 & 0.307 & 0.561  & \textbf{0.495} & 0.180 & \textbf{0.239} & 0.473 \\
     WebQA-LLaVA$_{1.5}$  & 0.572 & \textbf{0.262}    & 0.304    & 0.562 &0.456 &  0.184   &   0.237& \textbf{0.474}\\ 
     Med-MARVEL-LLaVA$_{1.5}$ & 0.581 & 0.260    & \textbf{0.311}  & \textbf{0.563} & 0.475   & \textbf{0.185 } & 0.236 &\textbf{0.474} \\  
     ClueWeb-LLaVA$_{1.6}$   & 0.601 & 0.252  & 0.303 & 0.558 & 0.492 & 0.178  & 0.237& 0.471  \\
    \bottomrule
\end{tabular}
}
\caption{Ablation study of FactMM-RAG including multimodal retrieval and backbone variation.}
\vspace{-0.5cm}
\label{tab:ablation}

\end{table*}

 % 0.408 & 0.182 & 0.238 & 0.471 \\
 % 0.406 & 0.183 & 0.241 & 0.471
\subsection{Ablation Study }
\label{subsec:ablation}

\textbf{Multimodal Retrieval.} Instead of relying on the multimodal foundation model to generate reports, we also evaluate the performance of the multimodal retrievers by directly encoding radiology images from the testing corpus and searching for the closest report from the training corpus for comparison with ground-truth reports.
Table \ref{tab:ablation} shows that our retriever also achieves the best factual retrieval performance compared to other baselines under this setting across two datasets. 
This demonstrates that training the multimodal retriever with mined factually-informed report pairs can enhance its radiology image understanding capabilities and directly align it with precise reports.
\\
\newline
\textbf{Backbone Variation.} We also investigate the impact of different retriever and foundation model backbones on radiology report generation in Table \ref{tab:ablation}.
We initialize our retriever model from two checkpoints: WebQA and ClueWeb in \citep{zhou2024marvel}.
We observe that the ClueWeb checkpoint provides a marginal gain compared to the WebQA checkpoint.
This can be attributed to the larger scale of the ClueWeb dataset used for pretraining.
We also utilize Med-MARVEL as our retriever backbone, which exhibits similar performance to other backbones after training.
This implies that even if our retriever is initialized with a backbone from a general domain, our factually-informed training strategy enables it to fully leverage medical knowledge and quickly adapt to the radiology-specific domain without degrading performance.

\subsection{Fact-aware Capability Control}
\label{subsec:threshold}
\begin{figure}[t]
    \centering
        \centering
        \includegraphics[width=0.48\textwidth]{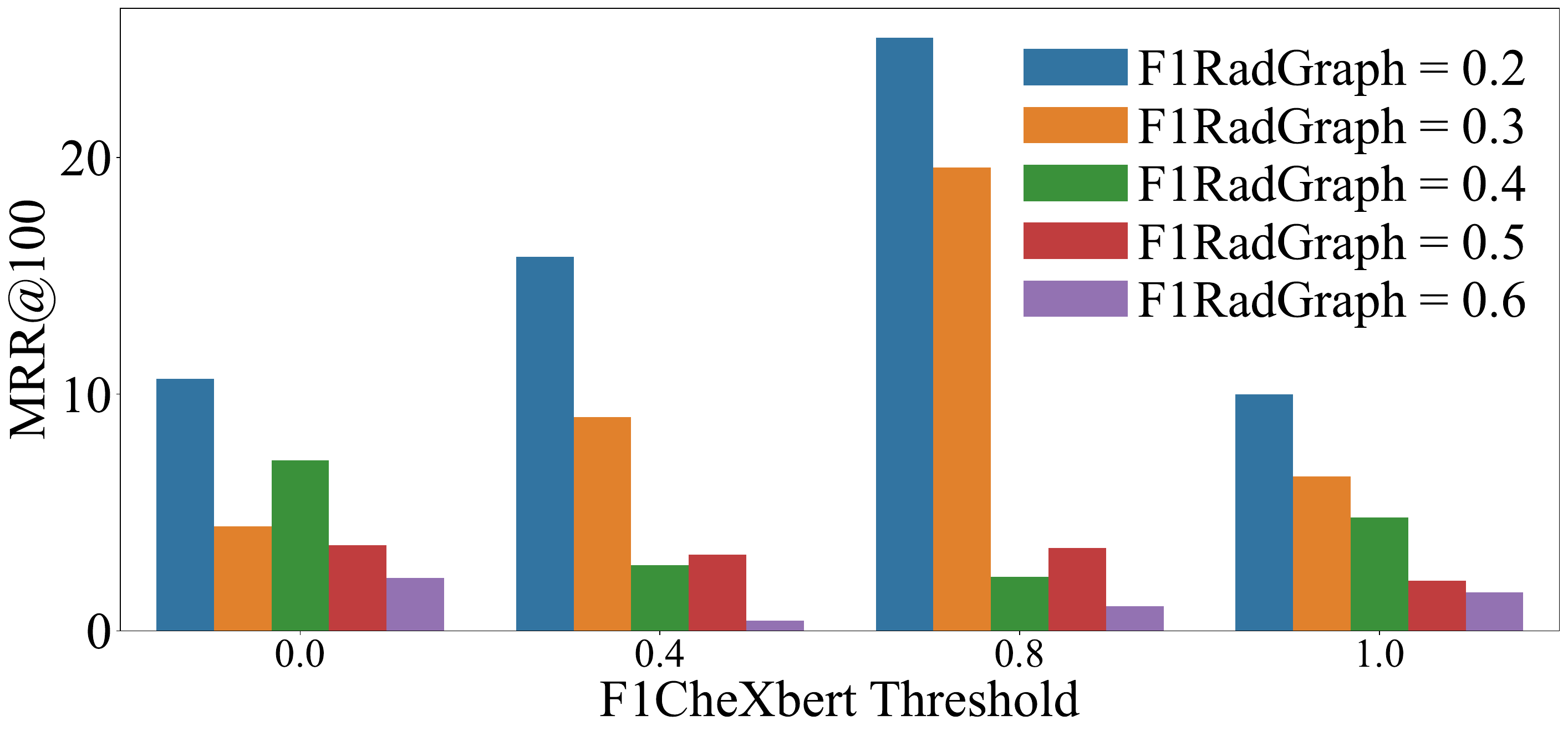}
        \vspace{-0.5cm}
   \caption{Retrieval evaluation of FactMM-RAG with different F1CheXbert and F1RadGraph thresholds. MRR calculates the mean reciprocal of rank at which the first relevant report that meets two factual similarity thresholds with query report is retrieved. }
    \label{fig:hyper_mrr}
\end{figure}

The factual similarity threshold in Equation \ref{eq:radgraph_similarity} plays a critical role in controlling the fact-awareness of our multimodal retriever.
We examine the performance of FactMM-RAG under different thresholds, as shown in Figure \ref{fig:hyperparameter_search}. 
Not only utilizing F1RadGraph thresholds, we also employ F1CheXbert to curate additional thresholds from the report's diagnostic labels to mine report pairs.\\
\newline
Under the same F1CheXbert threshold for mining report pairs, we observe that an increase in the F1RadGraph threshold correlates with an improvement in factual performance.
However, adopting stricter thresholds for identifying report pairs does not yield further improvements and reaches saturation.
After calculating the average number of report pairs per query, we find that high thresholds can exclude many relevant report pairs, as shown in Figure \ref{fig:hyper_mrr}. This exclusion results in the potential loss of factually useful pairs, thereby hindering the training of our multimodal retriever driven by additional factual medical knowledge.\\
\newline
Rather than relying on diagnostic labels from CheXbert to identify high-quality report pairs, Figure \ref{fig:chex0} demonstrates that the F1RadGraph threshold alone can also effectively mine factual report pairs for training our multimodal retriever.
As the F1RadGraph threshold increases, FactMM-RAG even matches the performance under high threshold settings in Figure \ref{fig:chex1.0}.
This signifies that employing our training strategy with curated factual query-report pairs still imposes useful supervision signals without relying on explicit diagnostic label guidance.
\begin{table*}[t]
    \centering

\resizebox{\textwidth}{!}{
    \begin{tabular}{c|p{6cm}|p{6cm}|p{6cm}}

        \hline
        \textbf{Radiology Image} & \textbf{Med-MARVEL} & \textbf{FactMM-RAG} & \textbf{Reference} \\
        \hline
    \raisebox{-\totalheight}{\includegraphics[width=5.5cm]{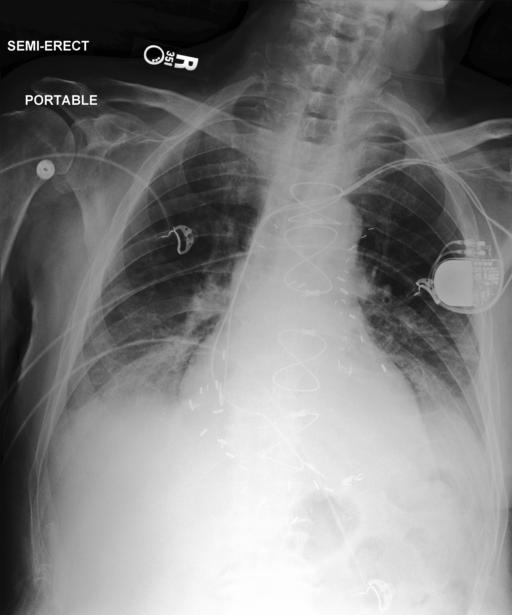}} & Single portable view of the chest. \textcolor{cyan}{\textit{There are bilateral pleural effusions,}} moderate on the left and small on the right. \textcolor{cyan}{\textit{There is also pulmonary vascular redistribution and hazy alveolar infiltrate. cardiac silhouette is enlarged}} but unchanged. Median sternotomy wires and mediastinal clips are again noted.& A left-sided pacemaker is in place with leads terminating in the right atrium and right ventricle.\textit{ \textcolor{orange}{The patient is status post median sternotomy and CABG.}} The heart is moderately enlarged. \textit{ \textcolor{cyan}{There is mild pulmonary edema. A small left pleural effusion is present.}} \textit{ \textcolor{red}{There is atelectasis at the left lung base.} }\textit{\textcolor{orange}{No pneumothorax is seen.}}& \textit{\textcolor{orange}{The patient is status post median sternotomy and CABG.}} Left-sided dual-chamber pacemaker is noted with leads terminating in right atrium and right ventricle, unchanged. Cardiomegaly is similar. \textit{\textcolor{cyan}{There is continued mild to moderate pulmonary edema}}, slightly improved compared to the prior exam. \textcolor{cyan}{\textit{Small layering bilateral pleural effusions}} also may be slightly decreased in the interval. \textit{ \textcolor{red}{Bibasilar airspace opacities likely reflect atelectasis.}} \textit{\textcolor{orange}{There is no pneumothorax.}} No acute osseous abnormalities are visualized.\\
        \hline
        F1RadGraph& 0.218&0.413 &   \\
        \hline
       CheXbert Observations & Cardiomegaly, Edema, Pleural Effusion & Cardiomegaly, Edema, Atelectasis, Pleural Effusion & Cardiomegaly, Edema, Atelectasis, Pleural Effusion  \\
        \hline
    \raisebox{-\totalheight}{\includegraphics[width=5.5cm]{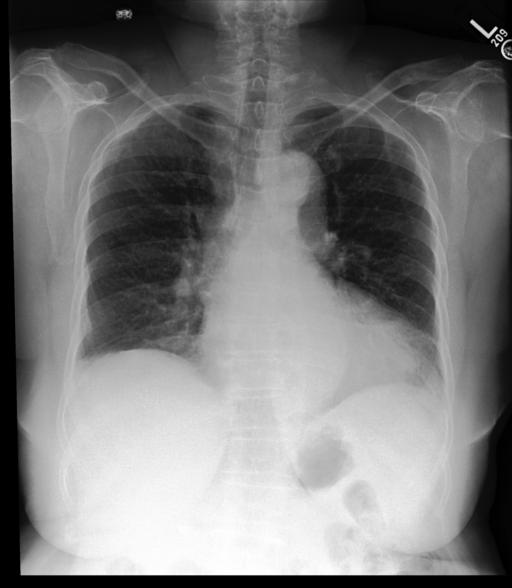}} &  \textcolor{cyan}{\textit{The heart is mildly enlarged. The aorta is mildly tortuous}.} The mediastinal and hilar contours appear unchanged. \textcolor{cyan}{\textit{There is no pleural effusion or pneumothorax.}} Streaky left basilar opacity suggests minor atelectasis. \textcolor{cyan}{\textit{There is no definite pleural effusion or pneumothorax.}} The bones appear demineralized. There is mild-to-moderate rightward convex curvature centered along the mid thoracic spine. & \textcolor{cyan}{\textit{Heart size is mildly enlarged. The aorta is tortuous.}} Mediastinal and hilar contours are otherwise unremarkable.\textcolor{orange}{\textit{ Pulmonary vasculature is normal.}} Linear opacities in the left lower lobe are compatible with subsegmental atelectasis.
    \textcolor{cyan}{\textit{ No focal consolidation, pleural effusion or pneumothorax is present.}} \textcolor{orange}{\textit{There are no acute osseous abnormalities. }} & \textcolor{cyan}{\textit{Moderate enlargement of the cardiac silhouette }}with a left ventricular predominance is unchanged. \textcolor{cyan}{\textit{The aorta remains tortuous}}, and the hilar contours are stable. \textcolor{orange}{\textit{Pulmonary vascularity is not engorged.}} There is minimal atelectasis within the lung bases, but \textcolor{cyan}{\textit{no focal consolidation is present. No pleural effusion or pneumothorax is identified}.} \textcolor{orange}{\textit{There are no acute osseous abnormalities. }}\\
        \hline
        F1RadGraph& 0.333&0.526 &   \\
        \hline
       CheXbert Observations & Cardiomegaly, Atelectasis & Cardiomegaly, Atelectasis& Cardiomegaly, Atelectasis \\
        \hline
    \end{tabular}
}
            \caption{Case study on generated reports from MIMIC-CXR. \textcolor{cyan}{Cyan} text indicates radiological consistency with the ground-truth report. \textcolor{orange}{Orange} text highlights extra accurate details provided by FactMM-RAG compared to Med-MARVEL. \textcolor{red}{Red} text denotes observations missing in Med-MARVEL. }
        \label{tab:case_study}
\end{table*}
\subsection{Fact-aware Capability Propagation }
\label{subsec:propa}
\begin{figure}[t]
    \centering
    \hspace{-0.5cm}
     \begin{subfigure}[t]{0.25\textwidth}
        \centering
        \includegraphics[width=\textwidth]{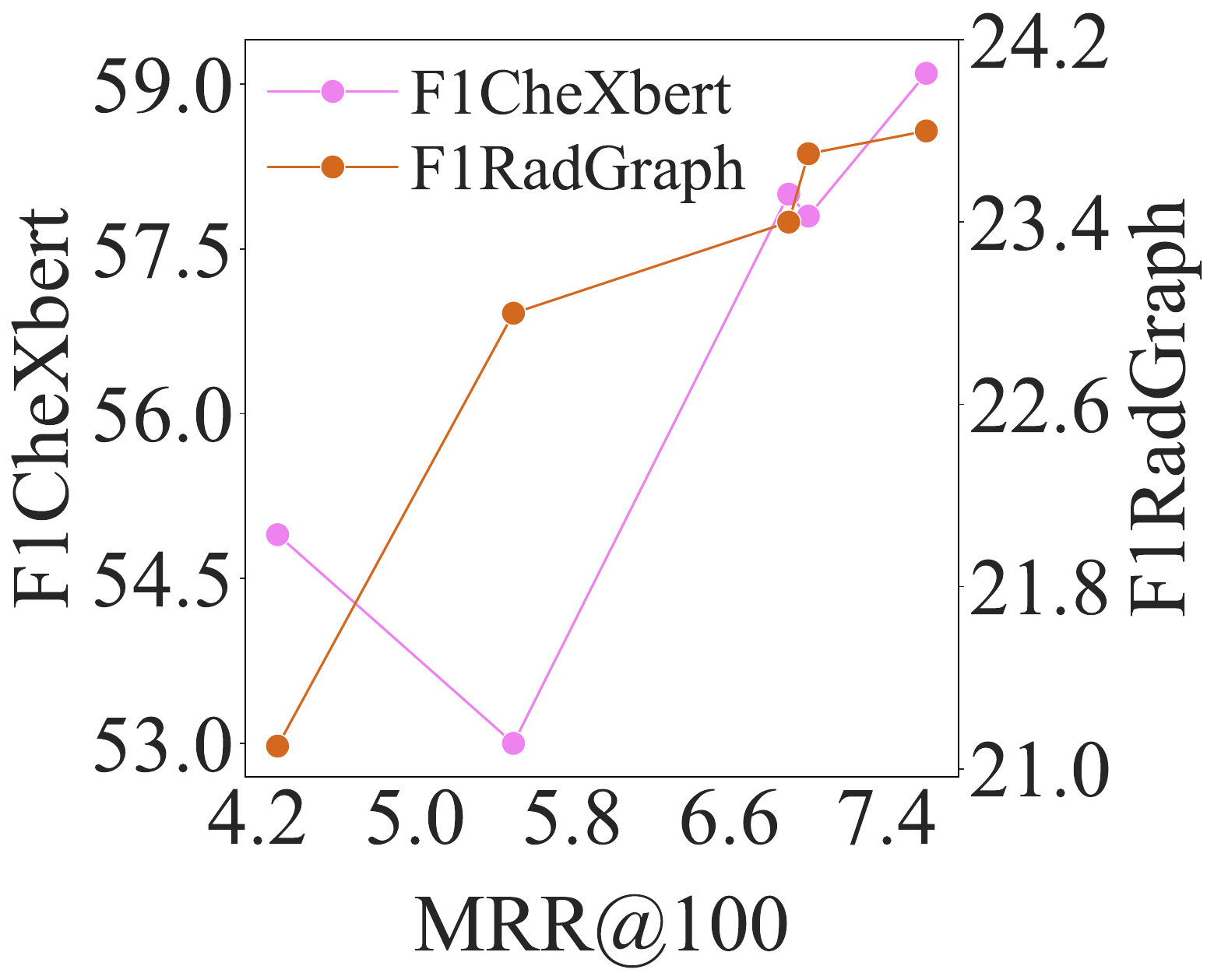}
        \caption{Multimodal Retrieval}
         \label{fig:prop_mr}
    \end{subfigure}~
        \hspace{-0.2cm}
    \begin{subfigure}[t]{0.25\textwidth}
        \centering
        \includegraphics[width=\textwidth]{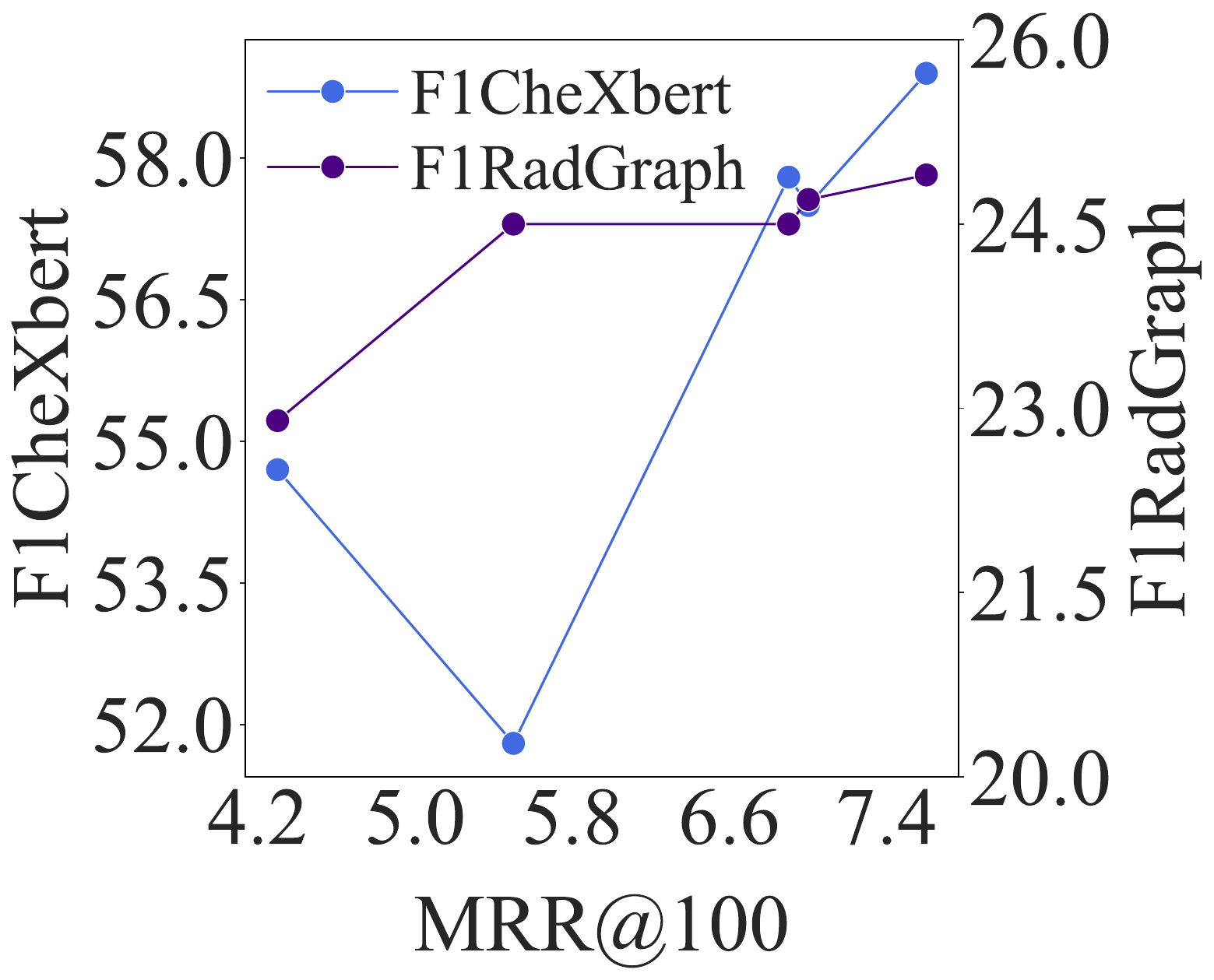}
        \caption{Multimodal RAG}
        \label{fig:prop_rag}
    \end{subfigure}
   \caption{Analysis of fact-aware capability propagation. The $x$-axis MRR measures the retriever's performance on retrieving factually relevant reports.}
    \label{fig:prop}
\end{figure}
To further understand the benefits of our retriever for the foundation model, we explore the effective propagation of fact-aware capabilities from the retriever to the foundation model.
To demonstrate this behavior, we use the mined factual report pairs as reference reports for the query report.
We then use the retrieval metric Mean Reciprocal Rank (MRR) as an intermediate evaluation, shown in Figure \ref{fig:prop}.
%
% to evaluate whether training the retriever with mined report pairs can enhance its fact-aware capability.
%
From the plot, we observe that as training progresses, the retrieval metric increases alongside two clinical metrics.
This factually-oriented upward trend in our retriever's performance in Figure \ref{fig:prop_mr} is also reflected in the foundation model's performance in Figure \ref{fig:prop_rag}.
This indicates that employing a factually-informed reference report selection strategy to train our multimodal retriever can also enhance the foundation model's ability to generate factually accurate radiology reports.
\subsection{Case Study}
\label{subsec:case}
In this section, we present two examples from MIMIC-CXR to qualitatively analyze our retriever's fact-aware capability, as illustrated in Table \ref{tab:case_study}.
In the first example, we observe that FactMM-RAG provides symptom observations consistent with the ground-truth report and generates more accurate factual details compared to Med-MARVEL, e.g., ``post median sternotomy, atelectasis, not pneumothorax'';
In the second example, we further observe that although both retrievers generate reports with diagnostic labels matching the ground-truth report, FactMM-RAG provides additional details compared to Med-MARVEL, such as ``pulmonary vasculature is normal, no acute osseous abnormalities''.
These characteristics confirm that adopting our fact-aware retriever can assist multimodal foundation models in generating more accurate radiology reports.
We show retrieved reports from two samples in Appendix \ref{subsec:app_case}.

\section{Conclusion}
\label{sec:conclusion}

In this paper, we aim at improving radiology report generation by introducing a fact-informed medical multimodal retriever for retrieval-augmented generation. 
In particular, we utilize RadGraph to annotate chest radiograph reports and mine clinically-relevant pairs. 
We integrate factual information into a universal multimodal retriever, presenting FactMM-RAG, a fact-aware multimodal retrieval-augmented radiology report generation pipeline.
FactMM-RAG outperforms all state-of-the-art retrievers evaluated by factual correctness and textual coherence for final report generation in MIMIC-CXR and CheXpert datasets.
We further confirm the benefit of our multimodal retriever from the analysis of its fact-aware capability.
% 
% Given the pervasive applications of machine learning in clinical diagnoses using chest X-rays, we hope our factual-informed approach inspires further work in multimodal generative artificial intelligence in healthcare contexts.
\section{Limitations}
\label{sec:limitation}
Despite the strong performance of our FactMM-RAG pipeline, we acknowledge potential limitations of our proposed method. 
In particular, our work only emphasizes chest radiology domains.
It also worth exploring our retrieval-augmented factual report generation pipeline in broader medical domains, such as brain scan or histology datasets.\\
% may not have available factual-correctness metrics of sufficient data to build our proposed pipeline.\\
%
\newline
Another concern lies in the chosen evaluation metrics, F1RadGraph and F1CheXbert. 
F1CheXbert reflects high-level observational accuracy, while F1RadGraph assesses the correctness of radiology entities and clinical relationships. However, other radiologically-specific metrics, such as report conciseness and clarity, should also be considered \citep{sureka2014seven}.
Ideally, we should incorporate methods of evaluation directly aligned with human evaluations or involve domain expertise itself in our pair-mining and final evaluation procedure. 
Moreover, it is worth performing a long-tail evaluation by leveraging more fine-grained ground-truth label annotations \citep{holste2023cxr}. 

\section{Ethics Considerations}
\label{sec:ethics}
A key ethical consideration in our work is the use of de-identified, credentialed medical data. In particular, the responsible usage policy of the MIMIC-CXR dataset strictly prohibits sharing access with third parties. To gain access to MIMIC-CXR, we completed a training course and signed a data use agreement, ensuring compliance with patient privacy regulations.

% Bibliography entries for the entire Anthology, followed by custom entries
%\bibliography{anthology,custom}
% Custom bibliography entries only
\bibliography{custom}
\appendix
\section{Appendix}
\label{sec:appendix}

\subsection{Retriever Training Procedure}
\label{subsec:app_ret}
To training our fact-aware multimodal retriever, we not only use mined factual report pairs as positive reports to the query image, but also incorporate the query image's corresponding report. 
Following \citep{inproceedings_ance,zhou2024marvel}, we also adopt modality-balanced hard negatives to train the retriever after in-batch negative training from the multimodal dense retrieval stage. 
We use AdamW \citep{loshchilov2019decoupled} as our optimizer and training epochs = 15, early stopping epoch = 5, batch size = 32, learning rate = 5e-6, and the temperature hyperparameter $\tau$ = 0.01. For our MARVEL backbone, we use T5-ANCE \citep{inproceedings_ance} as the text encoder and vision transformer \citep{dosovitskiy2021image} as the vision encoder. Models are trained using 1 NVIDIA RTX A6000 for 10 hours.

\subsection{RAG Finetuning Procedure}
\label{subsec:app_reg}
To create a RAG dataset for fine-tuning LLaVA, we search the nearest-neighbor document $d_{txt}^*$ for a query image $q_{img}$ using a retriever's embeddings. 
We filter out any results that involve retrieving a patient's own report, the same patient's other studies, or malformed reports in the training dataset (specified by being less than 5 characters). 
We apply the prompt templates in Figure \ref{fig:prompting}, and fine-tune LLaVA-1.5 for one epoch. Models are trained using 8x NVIDIA RTX A6000 for 4 hours, with epochs=1, learning rate=2e-5, global batch size=128, from vicuna-7b-v1.5 checkpoint. 
We save the checkpoint after one full pass of the training dataset for final evaluation. 
\definecolor{lasallegreen}{rgb}{0.03, 0.47, 0.19}
\begin{figure}[t]
    \textbf{Visual Question Answering}:
    \begin{tcolorbox}
    Generate a radiology report from this image: \color{lasallegreen}\texttt{<image>}\color{black}
    \end{tcolorbox}
    \textbf{Retrieval Augmented Generation}:
    \begin{tcolorbox}
    Here is a report of a related patient: "\color{lasallegreen}\texttt{<document>}\color{black}"\\
    Generate a radiology report from this image: \color{lasallegreen}\texttt{<image>}\color{black}
    \end{tcolorbox}
    \caption{Prompt templates for Visual Question Answering and Retrieval Augmented Generation}
    \label{fig:prompting}
\end{figure}
\begin{table*}[t]
    \centering

\resizebox{\textwidth}{!}{
    \begin{tabular}{c|p{6cm}|p{6cm}|p{6cm}}
        \hline
        \textbf{Radiology Image} &  \textbf{Med-Marvel}  & \textbf{FactMM-RAG} & \textbf{Reference} \\
        \hline
    \raisebox{-\totalheight}{\includegraphics[width=5.5cm]{Figures/case_study1.jpg}} & A single portable chest radiograph was obtained.\textcolor{cyan}{\textit{ Bilateral pleural effusions and mild atelectasis have increased.}} Cardiomegaly is unchanged. There is no consolidation or pneumothorax. Pacing leads, sternotomy wires, vascular clips, and abdominal surgical clips are unchanged. & No focal consolidation is identified. There is unchanged appearance of opacifications in the left lung base, \textcolor{cyan}{\textit{likely due to a combination of atelectasis and pleural effusion. There is a small right pleural effusion.}} \textcolor{red}{\textit{Mild pulmonary edema persists.}} The heart is moderately enlarged, but stable. Left sided pacemaker is seen with transvenous leads in the right atrium, right ventricle, and left ventricle. & The patient is status post median sternotomy and CABG. Left-sided dual-chamber pacemaker is noted with leads terminating in right atrium and right ventricle, unchanged. Cardiomegaly is similar. \textcolor{red}{\textit{There is continued mild to moderate pulmonary edema,}} slightly improved compared to the prior exam. \textcolor{cyan}{\textit{Small layering bilateral pleural effusions also may be slightly decreased in the interval. Bibasilar airspace opacities likely reflect atelectasis}.}There is no pneumothorax. No acute osseous abnormalities are visualized.\\
        \hline
        F1RadGraph& 0.274& 0.345&   \\
        \hline
       CheXbert Observations& Cardiomegaly,  Atelectasis, Pleural Effusion & Cardiomegaly, Edema, Atelectasis, Pleural Effusion & Cardiomegaly, Edema, Atelectasis, Pleural Effusion  \\
        \hline
    \raisebox{-\totalheight}{\includegraphics[width=5.5cm]{Figures/case_study2.jpg}} &\textcolor{cyan}{\textit{ The heart is mildly enlarged}} with a left ventricular configuration. There is mild-to-moderate unfolding of the thoracic aorta. The arch is partly calcified. The mediastinal and hilar contours appear unchanged. \textcolor{cyan}{\textit{There are streaky left basilar opacities suggesting minor atelectasis.}} A small eventration is noted along the anterior right hemidiaphragm. There is an air-fluid level in the stomach. Air-fluid levels are seen in the epigastric region. There is no evidence for free air. Cholecystectomy clips project over the right upper quadrant. Moderate degenerative changes are similar along the mid thoracic spine. & \textcolor{cyan}{\textit{Moderate enlargement of the cardiac silhouette is unchanged.}} \textcolor{orange}{\textit{The aorta remains tortuous. }}The mediastinal and \textcolor{orange}{\textit{hilar contours are normal.}} Pulmonary vasculature is normal. \textcolor{cyan}{\textit{Streaky atelectasis is noted }}in the left lower lobe. The right lung is clear. \textcolor{orange}{\textit{No focal consolidation, pleural effusion or pneumothorax is present.}} Multiple clips are noted within the left upper abdomen. & \textcolor{orange}{\textit{Moderate enlargement of the cardiac silhouette}} with a left ventricular predominance is unchanged. \textcolor{orange}{\textit{The aorta remains tortuous, and the hilar contours are stable.}} Pulmonary vascularity is not engorged. There is \textcolor{cyan}{\textit{ minimal atelectasis within the lung bases}}, but \textcolor{orange}{\textit{no focal consolidation is present. No pleural effusion or pneumothorax is identified.}} There are no acute osseous abnormalities.  \\
        \hline
        F1RadGraph& 0.197 & 0.621&   \\
        \hline
       CheXbert Observations& Cardiomegaly, Atelectasis &Cardiomegaly, Atelectasis & Cardiomegaly, Atelectasis \\
        \hline
    \end{tabular}
}
 \caption{Case study on retrieved reports  from MIMIC-CXR. \textcolor{cyan}{Cyan} text indicates radiological consistency with the ground-truth report. \textcolor{orange}{Orange} text highlights extra accurate details provided by FactMM-RAG compared to Med-MARVEL. \textcolor{red}{Red} text denotes observations missing in Med-MARVEL. }
    \label{tab:case_study2}
\end{table*}
\subsection{Evaluation Details}
\label{subsec:app_eva}
Here, we provide implementation details regarding the evaluation methodology.\\
\newline
\textbf{F1-RadGraph}. For F1-RadGraph score computation, we follow previous work (MIMIC-CXR-RRS) \footnote{\href{https://vilmedic.app/papers/acl2023/}{https://vilmedic.app/papers/acl2023/}} in employing $\textrm{RG}_{ER}$ as F1-Radgraph score computation on an instance level. Using the \texttt{radgraph} library implementation, this equates to utilizing \texttt{reward\_level="partial"}.\\
\newline
\textbf{F1-CheXbert}. F1-CheXbert score computation consists of the micro-averaged F1-score between 5 selected classes from the CheXbert labeler. Naturally, F1-CheXbert scores are only computable over entire datasets. For instance-level CheXbert scores (used for pair mining), we employ the proportion of equivalent predicted classes between a reference and predicted text sample. These instance-level F1-CheXbert scores can be computed using \texttt{np.sum(ref == hyp) / 5}, and take on values $\in\{0.0, 0.2, 0.4, 0.6, 0.8, 1.0\}$.\\
\newline
\textbf{CheXpert Hidden Test Set}. We use the 1000 hidden test reports from MIMIC-CXR-RRS and download the CheXpert images from Stanford AIMI Shared Datasets
\footnote{\href{https://stanfordaimi.azurewebsites.net/datasets/8cbd9ed4-2eb9-4565-affc-111cf4f7ebe2}{https://stanfordaimi.azurewebsites.net/datasets/8cbd9ed4-2eb9-4565-affc-111cf4f7ebe2}}.\\
% via the \texttt{azcopy} tool. 
\newline
\textbf{Oracle Retrieval}. Oracle Retrieval is performed via ground-truth access to a reference document's generated report. For training queries, this is always known, and an oracle retriever would obtain documents as $Oracle(q_{i})\doteq \underset{j \in \textrm{corpus}, j \neq i}{\arg\max}~s(q_i, d_j)$, where $s(q, d)$ is the sum of the F1-RadGraph and F1-CheXbert instance-wise scores. In practice, this results in retrieving samples with F1-CheXbert=$1.0$ and the largest F1-RadGraph score within the partition. Test-time retrieval performs the same operation, without the restriction of $j \neq i$ as self-retrieval is not possible due to the corpus being the training dataset.\\
\newline
\textbf{Oracle RAG}. Oracle-LLaVA is obtained by fine-tuning LLaVA under identical conditions, utilizing Oracle Retrieval for retrieving documents in the training and test set.

\subsection{Case study on Retrieved Reports}
\label{subsec:app_case}
We now conduct case study on the retrieved reports shown in Table \label{tab:case_study2}.
We show that our FactMM-RAG captures most of the factual details in retrieved reports compared to the ground-truth reports.
Thus, the factual correctness of our retriever can be propagated to the multimodal foundation models effectively.
However, the reports retrieved from Med-MARVEL contain erroneous information, which negatively impacts the report generation by multimodal foundation models.

\end{document}